# Multichannel Distributed Local Pattern for Content Based Indexing and Retrieval

Sonakshi Mathur, Mallika Chaudhary, Hemant Verma, Murari Mandal, S. K. Vipparthi
CSE Department
MNIT Jaipur
mathursonakshi@gmail.com, mallika454@gmail.com, hemantv133@gmail.com, murari023@gmail.com, skvipparthi@mnit.ac.in

Subrahmanyam Murala
Department of Electrical Engineering
IIT Ropar
subbumurala@iitrpr.ac.in

*Abstract*— A novel color feature descriptor, Multichannel Distributed Local Pattern (MDLP) is proposed in this manuscript. The MDLP combines the salient features of both local binary and local mesh patterns in the neighborhood. The multi-distance information computed by the MDLP aids in robust extraction of the texture arrangement. Further, MDLP features are extracted for each color channel of an image. The retrieval performance of the MDLP is evaluated on the three benchmark datasets for CBIR, namely Corel-5000, Corel-10000 and MIT-Color Vistex respectively. The proposed technique attains substantial improvement as compared to other state-of-the-art feature descriptors in terms of various evaluation parameters such as ARP and ARR on the respective databases.

## I. Introduction

Rapid advances in imaging technology has given rise to an exponential increase in the size of digital libraries. Further, the readily available high-definition digital cameras are producing better quality images which are easily accessible on the internet. The manual handling of these image databases and image annotation is quite a time consuming and cumbersome task. For effective and efficient utilization of these visual data, seamless indexing and retrieval is an essential task. Also, the information overload caused by vast availability of visual information on internet is quite alarming. The automatic search technique called Content Based Image Retrieval (CBIR) was proposed to address these above-mentioned problems. CBIR retrieves images based on features like shape, colour, texture, spatial layout, etc. which are used to represent and index the image database rather than metadata such as tags or keywords. The feature extraction is very crucial for CBIR based techniques and its effectiveness is largely dependent upon the feature descriptor that has been chosen for an image. Feature descriptors play a crucial role in CBIR and numerous other applications like object recognition, biometrics, facial expression recognition, object tracking, medical imaging etc. As a result, different varieties of feature descriptors have been proposed in the literature. Early works in CBIR included utilizing the Gabor transform (GT) for extracting the texture information from the images. Manjunath and Ma [1] used the mean and standard deviation data from Gabor transformed image and collected information from six orientations and four scales. Han et al. [2] and Chall et al. [3] further incorporated the scale-invariance and rotation invariance properties by making modifications in the Gabor filters for image retrieval. Wavelet based approaches improved upon the previously proposed image retrieval systems. Wavelet correlogram and Gabor Wavelet Correlogram (GWC) based experiments were carried out by Moghaddam et al. [4]. Murala et al. [5] combined the wavelet and Rotated Wavelet Correlogram (RWC) to increase the retrieval performance. Literature surveys on CBIR and image feature descriptors is available in [6-8].

Colour and texture features are also significant features which are frequently used in the field of CBIR. Colour histograms, introduced by Swain and Ballar [9] has been frequently used in many image retrievals based applications. Stricker and M. Oreng [10] used a fusion of initial 3 central moments - mean, standard deviation and skewness of each colour channel. Huang et al. [11] characterized the pixel colour distribution and the spatial correlation of two colour channels by introducing a novel colour feature named, colour correlogram.

In the past few years, the local pattern descriptors have become very popular for feature extraction in number of applications like object classification, face recognition, texture image retrieval, biomedical image retrieval, facial expression recognition, retinopathy image classification, palm print recognition, object detection and tracking etc. Generally, a local descriptor extracts the features from local neighbourhood regions. Ojala et al. [12] proposed the first local pattern called Local Binary Pattern (LBP) for describing the texture of an image in terms of the relationship between a reference pixel and the surrounding neighbours in the local neighbourhood. LBP and its variants can quickly extract the texture features as compared to other traditional methods and have strong discriminative capability. Illumination change invariance property was incorporated in LTP (Local Ternary Patterns) [13], whereas relationship between the neighbourhood pixels were extracted by using LMeP (Local Mesh Patterns) [14].

The local descriptors based feature extraction techniques have shown significant performance improvement in the field of image retrieval [15-18] as well as other application areas. Vipparthi et al. [15] developed directional local motif XOR patterns for image retrieval applications. They further designed a colour based retrieval system using Color directional local quinary patterns [16]. In [17], a multi-joint histogram based modelling technique was proposed. Vipparthi et al. [18] incorporated edge information in their descriptor through LDMaMEP and applied it in face recognition and image retrieval applications. Murala et al. [19-22] has also achieved some significant performance improvement through their proposed descriptors like Local Tetra Patterns (LTrP), Local Maximum Edge Binary Pattern and Spherical Symmetric 3D Local Ternary Patterns. In [23], a block based LBP was proposed for local patch based texture extraction. Directional features and center-symmetric LBP features were extracted by [24] and [25] respectively.

The various LBP variants proposed in the literature generally try to address the challenges that occur due to variations in illumination, noise, etc. in numerous applications. The LBP extracts the relationship between a pixel and the pixels around it. On the other hand, the LMeP extracts the relationship amongst the neighbours of a reference pixel. In this manuscript, we propose a robust local descriptor named Multichannel Distributed Local Pattern (MDLP), which integrates the relevant properties of both the LBP and LMeP. The MDLP captures information at various distances for the unichrome colour channels. The MDLP performs well in CBIR applications and the robustness of the same is analysed by performing experiments on three benchmark datasets, Corel-5k [26], Corel-10k [26] and MIT-color VisTex [27].

In the remaining part of this paper, we first present the related work in Section II. In the Section III, the proposed MDLP method, CBIR framework and query matching techniques are discussed. Evaluation parameters and the experimental results have been presented in Section IV. The paper is concluded in Section V.

## II. RELATED WORK

### A. Local Binary Patterns

Ojala et al. [12] introduced the concept of LBP for describing the texture of an image in terms of the relationship between a pixel and the nearby neighbours in the local neighbourhood. For a grayscale image $J$ of size $X \times Y$, let $J(m,n)$ be a location coordinate where $m \in [1, X]$ and $n \in [1, Y]$. The LBP descriptor as depicted in Fig. 1, can be computed based on Eq. (1).

$$LBP_{Nb,R}(m,n) = \sum_{i=0}^{Nb-1} 2^i \times S(g_i - g_c) \qquad (1)$$

where $Nb$ denotes the number of neighborhood pixels, $R$ denotes the radius, $g_c$ and $g_i$ denotes the intensity level of the central and neighborhood pixel. The sign function $S(\cdot)$ is computed through Eq. (2).

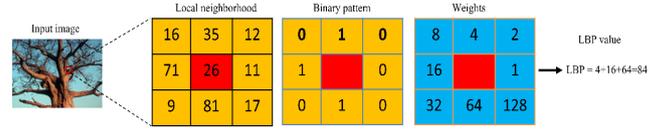

**Fig. 1** Local Binary Pattern

$$S(x) = \begin{cases} 1 & x \geq 0 \\ 0 & \text{otherwise} \end{cases} \qquad (2)$$

### B. Local Mesh Patterns

Murala et al. [14] proposed LMeP based on relationship between surrounding neighbours for a given pixel. The LMeP descriptor can be computed using Eq. (3) – Eq. (5).

$$LMeP_{Nb,R}^d(m,n) = \sum_{i=1}^{Nb} 2^{i-1} \times S(g_{\beta|R} - g_{i|R}) \qquad (3)$$

$$\beta = 1 + \text{mod}((i + Nb + d - 1), Nb) \qquad (4)$$

$$\forall d = 1, 2, ..., (Nb/2) \qquad (5)$$

where $d$ represents LMeP index and $R$ represents the radius of the neighboring pixels in consideration.

## III. MULTICHANNLE DISTRIBUTED LOCAL PATTERN (MDLP)

### A. The Proposed Feature Descriptor

The proposed MDLP descriptor inherits the salient properties of both LBP and LMeP, i.e., the relative change information between a pixel and its neighbors as well as between the neighbors in the local neighborhood. The MDLP is able to capture the texture information at various distances for the unichrome color channels. Thus, fusion of both the color and texture information are achieved for better performance. Let $J^t$ is the $t^{th}$ channel of an image $J$ of size $X, Y, C$, where $t \in [1, C]$ and $C$ is the total number of colour channels. Let the $Nb$ neighbours at radius $R$ of any pixel with location coordinates $J^t(m,n)$ for $m \in [1, X]$ and $n \in [1, Y]$ are defined as $J_t^u(m,n)$, where $u \in [1, Nb]$. Then MDLP for each color channel $t$ can be computed as shown below,

$$MDLP_{Nb,R}^{t,d}(m,n) = \{MDLP^1(m,n), MDLP^2(m,n)\} \qquad (6)$$

$$MDLP^1(m,n) = LBP_{Nb,R}^t(m,n) \qquad (7)$$

$$MDLP^2(m,n) = LMeP_{Nb,R}^{t,d}(m,n) \qquad (8)$$

where $d = 1, 2, 3, 4$ is the various distances at which the LMeP is computed. The LBP and LMeP can be computed through Eq. (1) and Eq. (3) respectively. The MDLP for $t^{th}$ color space contains one LBP and four LMeP patterns.

### B. MDLP Feature Vector

Two major components of pattern recognition applications are feature representation and matching/classification techniques. For delivering competitive performance, it is essential to ensure robust implementation of both parts. The MDLP histogram $HIST$ is generated as the feature vector for robust image feature representation where $HIST = \{HIST_t\}$, $t \in [1, C]$. Let $J(m,n)$ be a color image of size $X, Y, C$ with $W$ intensity levels, i.e., $I \in \{0,1,...,W-1\}$, where $(m,n)$ is the location coordinate for $m \in [1, X]$ and $n \in [1, Y]$. The MDLP histogram $HIST_t(\cdot)$ is computed as given in Eq. (9),

$$HIST_t(MDLP) = \sum_{m=1}^{X} \sum_{n=1}^{Y} \xi(MDLP_{Nb,R}^{t,d}(m,n), I) \quad (9)$$

$$\xi(u,v) = \begin{cases} 1, & u = v \\ 0, & otherwise \end{cases} \quad (10)$$

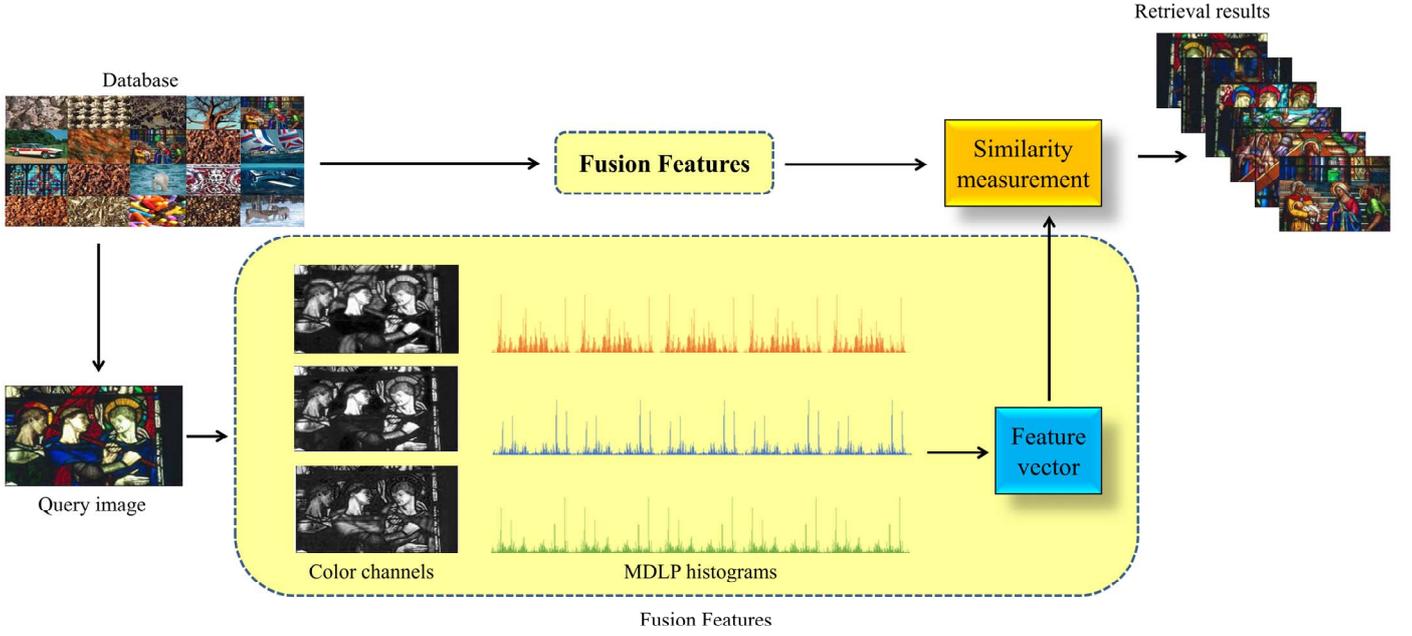

**Fig. 2** Proposed CBIR system framework

### C. The Proposed CBIR System Framework

The algorithm for the proposed CBIR system is presented in Algorithm 1.

---
**Algorithm 1.** Proposed CBIR System Framework
*Input:* Input Image
*Output:* Output Images
*Step 1:* Take the input query image
*Step 2:* MDLP feature vector computation.
*Step 3:* Query image matching with the database images.
*Step 4:* Retrieval of response images.

---

### D. Query Matching

The performance of a CBIR system also depends on the choice of similarity metrics. For a query image $Q$, the feature vector can be represented as $F_Q$ and all the database images from $DB$ can be represented by the feature vector $F_{DB_j}$ where $j = 1, 2, ....., |DB|$. For retrieval response, the $n$ top matched images which resemble the query image are selected from the database. The similarity matching is performed by using $d_1$ – *Distance* as shown in Eq. (11)

$d_1$ - *Distance:*

$$D_{d1}(Q, DB) = \sum_{i=1}^{|DB|} \frac{|F_{DB_i} - F_Q|}{|1 + F_{DB_i} + F_Q|} \quad (11)$$

The $F_{DB_i}$ is the feature vector of image $DB_i$ in the database and $F_Q$ is the feature vector of the query image. $|DB|$ is the database size.

## IV. EXPERIMENTAL RESULTS AND DISCUSSIONS

The retrieval performance of an image retrieval system is measured by Average Retrieval Rate (ARR) and Average Retrieval Precision (ARP). For a query image $Q$, ARR and ARP is defined as follows.

$$ARR = \frac{1}{|DB|} \sum_{i=1}^{|DB|} R(DB_i, n_r) \Big|_{n_r \leq N_t} \quad (12)$$

$$ARP = \frac{1}{|DB|} \sum_{i=1}^{|DB|} P(DB_i, n_r) \quad (13)$$

where $R$ and $P$ are the recall and precision computed using Eq. (14) and Eq. (15) respectively.

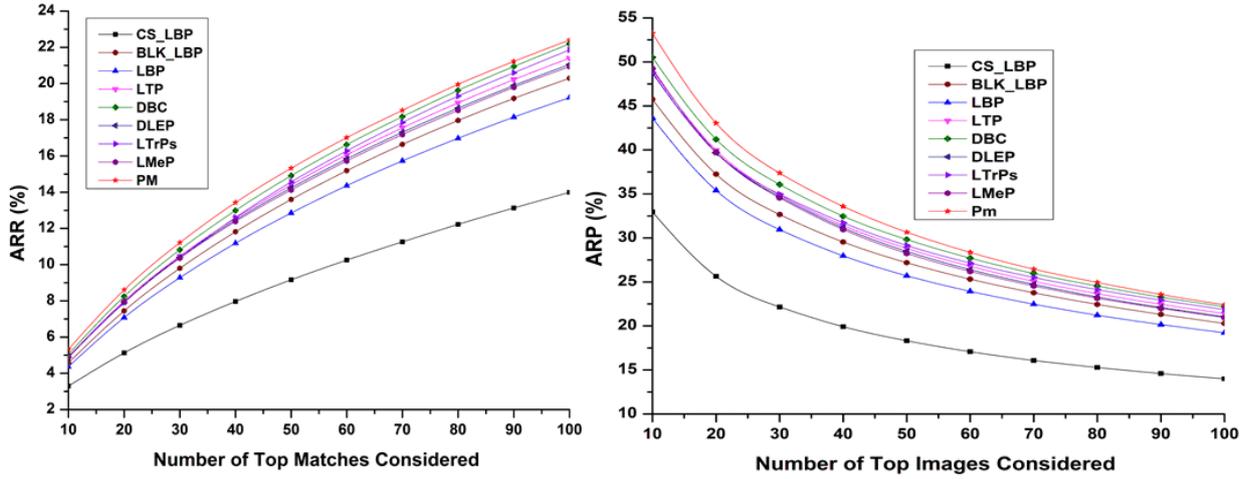

**Fig. 3** Comparative performance of the Proposed Method and other recent approaches in terms of ARR and ARP on Corel-5k dataset

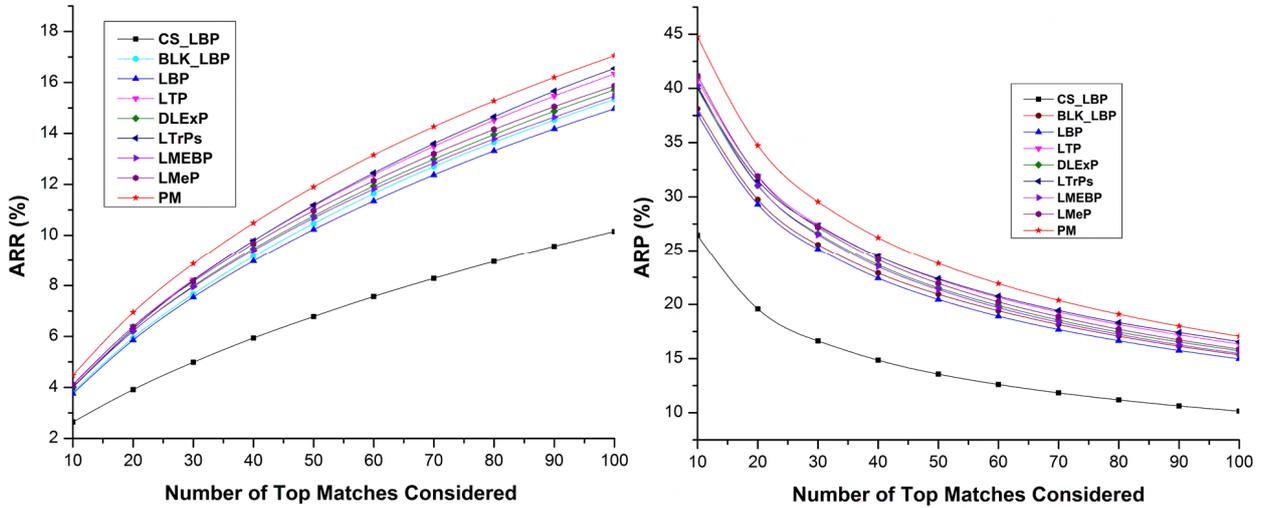

**Fig. 4** Comparative performance of the Proposed Method and other recent approaches in terms of ARR and ARP on Corel-10k dataset

$$R(Q, n_r) = \frac{1}{N_t} \sum_{i=1}^{|DB|} | \delta(\varphi(DB_i), \varphi(Q)) | Rank(DB_i, Q) \leq n_r | \quad (14)$$

where $n_r$ is the number of retrieved images and $N_t$ is the number of top matches for the retrieval system.

$$P(Q, n_r) = \frac{1}{n_r} \sum_{i=1}^{|DB|} | \delta(\varphi(DB_i), \varphi(Q)) | Rank(DB_i, Q) \leq n_r | \quad (15)$$

where $\varphi(p)$ is the category of $p$, $Rank(DB_i, Q)$ returns the rank of image $DB_i$ (for the query image $Q$) among all images of $DB$. The $\delta(\cdot,\cdot)$ is computed using Eq. (16)

$$\delta(\varphi(DB_i), \varphi(Q)) = \begin{cases} 1, & \varphi(DB_i) = \varphi(Q) \\ 0, & otherwise \end{cases} \quad (16)$$

### A. Experiment #1

The first experiment is carried out on Corel-5k [26] database. The database is made up of 5000 images each belonging to any one of the 50 categories. Each category consists of 100 images. For performance evaluation, both ARR and ARP were calculated. The proposed method, MDLP, achieves better ARP and ARR compared to several local descriptors based approaches such as LBP, BLK__LBP, LTP, CS_LBP, DBC, DLeP, LTrPs and LMeP. The comparative measures are shown in Fig. 3(a) and 3(b). In Fig. 6(a), the top 10 responses for a sample query image on Corel-5k database is depicted.

### B. Experiment #2

The second experiment is carried out on Corel-10k [26] database. The database is made up of 5000 images each belonging to any one of the 50 categories. Each category consists of 100 images. For performance evaluation, both AAR and ARP were calculated. The proposed method, MDLP, achieves better ARP and ARR compared to recent local descriptor based approaches such as LBP, BLK__LBP, LTP, DLExP, LTrPs, CS_LBP, LMEBP and LMeP. The comparative measures are shown in Fig. 4(a) and 4(b). In Fig. 6(b), the top 10 responses for a sample query image on Corel-10k database is depicted.

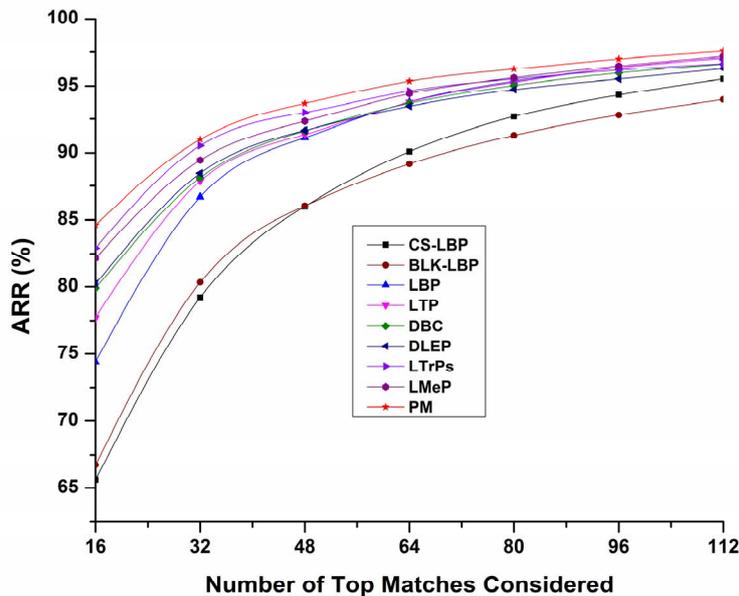

**Fig. 5** Comparative performance of the Proposed Method and other recent approaches in terms of ARR on MIT-color VisTex dataset

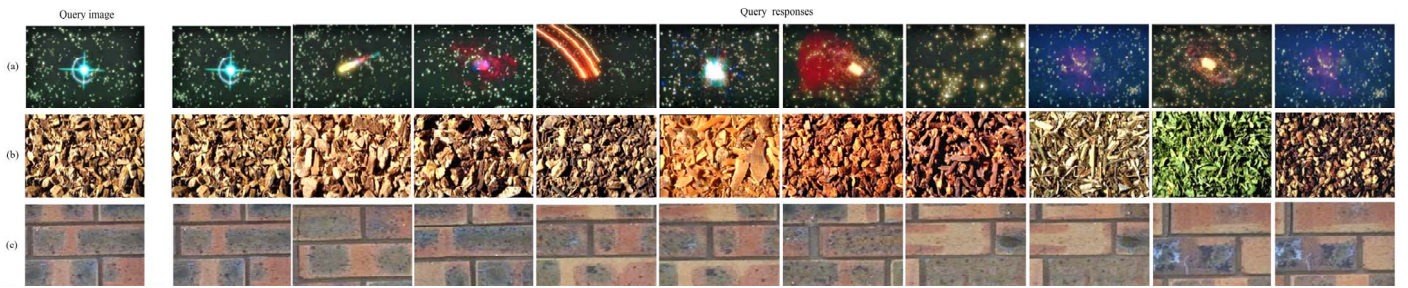

**Fig. 6** Top 10 retrieved image using the PM over (a) Corel-5k, (b) Corel-10k and (c) MIT-color VisTex datasets

### C. Experiment #3

In the last experiment, the color texture database MIT-color Vistex [27] is used to validate the performance of the proposed method. The database has 40 different color texture images of size $512 \times 512$. The textures are further divided into $128 \times 128$ sub images which are non-overlapping. For performance evaluation, both ARR and ARP were calculated. The proposed method, MDLP, achieves better ARP and ARR compared many recent approaches such as LBP, BLK__LBP, LTP, DBC, DLEP, LTrPs, CS_LBP and LMeP. In Fig. 6(c), the top 10 responses for a sample query image on MIT-color Vistex database is depicted.

### V. CONCLUSION

In this manuscript, a color and texture fusion based feature descriptor namely MDLP is proposed for image retrieval and analysis. MDLP combines the properties of LBP and LMeP to extract the texture features from all the color channels. The effective integration of color and texture information enhanced the robustness of MDLP. To evaluate the proposed method, experiments have been conducted over 3 datasets, Corel-5000, Corel-10,000 and MIT-Color VisTex which are publicly available. The MDLP descriptor shows a significant improvement of 4.02% and 9.63% for ARP (Corel-5000), 4.02% and 9.63% for ARP (Corel-10,000) and 2.46% and 10.22% for ARR (MIT-color) as compared to LMeP and LBP respectively.